\documentclass[12pt]{article}
\usepackage{geometry}                % See geometry.pdf to learn the layout options. There are lots.
\usepackage{graphicx}
\usepackage{amssymb}
\usepackage{amsbsy}
\newcommand{\Figure}[4]{%
\begin{figure}%[!t]
\centerline{\includegraphics#3{graphics/#2.pdf}}
\caption{#4}
\label{#1}
\end{figure}
}

\title{Topographic Representation\\
for Quantum Machine Learning}
\author{Bruce MacLennan\\
Department of Electrical Engineering and Computer Science\\
University of Tennessee, Knoxville}
\newcommand{\Omit}[1]{}
\def\bn{\begin{equation}}
\def\en{\end{equation}}
\def\bea{\begin{eqnarray*}}
\def\eea{\end{eqnarray*}}
\def\ifarray{\left\{ \begin{array}{ll}}
\def\endifarray{\end{array}\right.}

\def\cross{\mbox{\boldmath $\times$}}
\def\to{\rightarrow}
\def\onto{\twoheadrightarrow} %AMS sym
\def\oneone{\leftrightarrow}

\def\TP{\otimes}		% tensor product
\def\bigTP{\bigotimes}		% tensor product
\def\eqdf{\stackrel{{\rm def}}{=}}

\def\Complex{{\mathbb C}}

\def\Lin{{\cal L}}
\def\Eltwo{{\cal L}^2}

\def\Hilb{{\cal H}}
\def\arbdom{\Omega}
\def\altdom{\Omega'}
\def\intdom{\int_\arbdom}
\def\intalt{\int_{\altdom}}

\newcommand{\abs}[1]{| #1 |}

\newcommand{\norm}[1]{\| #1 \|}

\newcommand{\conjtrans}[1]{#1^{\dagger}}
\def\degree{^\circ}
\newcommand{\Bra}[1]{\langle #1 |}
\newcommand{\Ket}[1]{| #1 \rangle}
\newcommand{\BraKet}[2]{\langle #1 \mid #2 \rangle}

\newcommand{\KetBra}[2]{\Ket{#1}\Bra{#2}}

\def\SR2{\sqrt{2}}		% sqrt(2)
\def\RSR2{\frac{1}{\SR2}}	% 1 / sqrt(2)
\def\HR3{\frac{\sqrt{3}}{2}}	% sqrt(3) / 2
\def\Zero{{\bf 0}}

\def\Two{{\bf 2}}
\def\rv{{\bf r}}
\def\uv{{\bf u}}
\def\vv{{\bf v}}
\def\xv{{\bf x}}
\def\yv{{\bf y}}

\newtheorem{prop}{Proposition}

\newenvironment{prf}{\begin{trivlist} \item[{\bf ~Proof}:]}%
{\begin{flushright} $ \Box $ \end{flushright} \end{trivlist} }

\def\CNOT{{\rm CNOT}}

\def\CCNOT{{\rm CCNOT}}
\def\CSWAP{{\rm CSWAP}}
\def\I{{\rm I}}
\def\X{{\rm X}}

\def\H{{\rm H}}
\def\piby8{T}

\def\dif{{\rm d}}
\newcommand{\RSRx}[1]{\frac{1}{\sqrt{#1}}}
\def\map{\mu}

\def\sv{{\bf s}}
\def\wv{{\bf w}}
\def\inval{\xv}
\def\invalsp{{\cal X}}
\def\outval{\yv}

\def\loc{\rv}
\def\outloc{\sv}
\def\altloc{\sv}
\def\gmap{T}
\def\gmapbij{M}
\def\gmapnull{N}
\def\nrmap{S}
\def\Rmap{R}
\def\Qmap{Q}
\def\Pmap{P}
\def\Umap{U}
\def\Im{\mathop{\rm Im}}
\def\HilbAnc{\Hilb_{\rm C}}
\def\HilbGarb{\Hilb_{\rm G}}
\def\exinbasis{w}
\def\exoutbasis{v}
\def\Exinbasis{\omega}
\def\Exoutbasis{\upsilon}
\def\nrbasis{z}
\def\orcomp{^\perp}
\def\nullity{{n_o}}
\def\nullsp{{\cal N}}
\def\nonnbasis{\uv}
\def\nullbasis{\vv}
\def\nullbasism{\wv}

\def\nonnpart{\mu}
\def\nullpart{\nu}
\def\premult{n}
\def\csize{s}
\def\dsize{t}
\newcommand{\preimg}[2]{#1^{-1}\{#2\}}
\def\RSRni{\RSRx{n_i}}
\def\garbage{\gamma}

\def\nonrange{{m_{\rm nr}}}
\def\singval{{n_{\rm b}}}
\def\mulvalin{{n_{\rm n}}}
\def\mulvalout{{m_{\rm n}}}
\def\Unonrange{U_{\rm nr}}
\def\Usingval{U_{\rm b}}
\def\Umulval{U_{\rm n}}
\def\Umulpart{V}
\def\garbage{\gamma}
\newcommand{\OR}[1]{{\rm OR}_{#1}}
\newcommand{\qcomp}[1]{#1'} % quadratic complement

\newcommand{\zket}[1]{\Ket{0}_{#1}}
\newcommand{\zbra}[1]{\Bra{0}_{#1}}
\newcommand{\oket}[1]{\Ket{1}_{#1}}
\newcommand{\obra}[1]{\Bra{1}_{#1}}
\newcommand{\qargsin}[1]{[#1]}
\newcommand{\qargsout}[1]{[#1]\leftarrow}
\def\Sum{{\rm sum}}
\def\UOP{U_{{\rm OP}}}

\newcommand{\singket}[1]{\Ket{\{#1\}}}

\def\Udemux{U_{\rm demux}}

\begin{document}
\maketitle
\section{Introduction}
This paper proposes a brain-inspired approach to quantum machine learning with the goal of circumventing many of the complications of other approaches.
The fact that quantum processes are unitary presents both opportunities and challenges.
A principal opportunity is that a large number of computations can be carried out in parallel in linear superposition, that is, with quantum parallelism.
There are of course many technical challenges in quantum computing, but the principal theoretical challenge in quantum machine learning is the fact that quantum processes are linear whereas most approaches to machine learning depend crucially on nonlinear operations. Artificial neural networks, in particular, require a nonlinear activation function, such as a logistic sigmoid function, for nontrivial machine learning. 
Fortunately, the situation is not hopeless, for we know that nonlinear processes can be embedded in unitary processes, as is familiar from the circuit model of quantum computation.

Despite the complications of quantum machine learning, it presents a tantalizing approach to implementing large-scale machine learning in a post-Moore’s law technological era. However, there are many approaches to machine learning and several approaches to quantum computation (e.g., circuit model, adiabatic), and it is not obvious which combinations are most likely of success. Schuld, Sinayskiy, and Petruccione \cite{Schuld-QQNN,Schuld-IQML} provide useful recent reviews of some approaches to quantum neural networks and quantum machine learning more generally. They conclude that none of the proposals for quantum neural networks succeed in combining the advantages of artificial neural networks and quantum computing. They argue that open quantum systems, which involve dissipative interactions with the environment, are the most promising approach. Moreover, few of the proposals include an actual theory of quantum machine learning.

This paper explores an approach to the quantum implementation of machine learning involving nonlinear functions operating on information represented topographically, as is common in neural cortex.
However, there are many approaches to neurocomputing, that is, to brain-inspired computing, some of which may be more amenable to quantum implementation than others, and so we must say a few words about the alternatives. For the brain may be modeled at many different levels, and models at each of these levels may provide a basis for machine learning. For example, in recent years spiking neural network models have become popular once again, largely due to their power advantages when implemented in hardware 
\cite{Furber-SpiNNaker,Neftci2013}. 
Such models mimic the ``all or none'' action potential generation by biological neurons without addressing the detailed dynamics of the action potential. On the other hand, most contemporary applications of artificial neural networks, including those used in deep learning, use a higher level, rate-based model. That is, the real values passed between the units (neuron analogs) represent the rate of neural spiking rather than individual spikes. It has been argued that this is the appropriate level for modeling neural information processing, since there are many stochastic effects on the generation and reception of action potentials, and because the fundamental units of information processing are microcolumns comprising about 100 neurons \cite[ch.\ 2]{OReilly-CCN}. Therefore it is most fruitful to view neural computation as a species of massively parallel analog computation. Since quantum computation makes essential use of complex-valued probability amplitudes, it is also fruitful to treat it as a species of analog computation, and so analog information representation provides one point of contact between quantum computation and artificial neural networks \cite{FCFQIC}.

\section{Topographic Representation in the Brain}
Another respect in which information processing in the brain differs from most artificial neural network models is that biological neural networks are spatially organized, with connectivity dependent on spatial organization. Although artificial neural networks are typically organized in layers, there is generally no spatial relationship among the neurons in each layer;\footnote{
They are numerically indexed, of course, but interchangeable in terms of their pattern of connections before learning.}
the exceptions are convolutional neural networks, which were in fact inspired by the organization of sensory cortex.

One of the most common spatial information representations used by the brain is the \emph{topographic representation} or \emph{computational map}.
In such representations, distinct points $\inval$ in some abstract space $\invalsp$ are mapped systematically to physical locations $\loc = \map(\inval)$ in a two-dimensional region of cortex;
that is, spatial relationships among the neurons are correlated with topological relationships in the abstract space. These maps are especially common in motor areas \cite{Morasso-Sanguineti-SOCMMC} and sensory areas \cite[ch. 6]{OReilly-CCN}. 
For example, \emph{tonotopic} maps have the neurons that respond to different pitches that are arranged in order by pitch.
\emph{Retinotopic} maps have a spatial organization that mirrors the organization of the retina and visual field.
Neurons in primary visual cortex that respond to edges are arranged systematically according to the orientation of the edges.
There are many other examples throughout the brain, and this is perhaps the single most common information representation used by the brain.

In these topographic maps, a particular value $\inval$ is represented by an activity peak in the corresponding cortical location $\map(\inval)$.
The strength of the activity reflects the value's degree of presence or probability.
Moreover, multiple simultaneous values, with differing relative strengths or probabilities, are represented by multiple simultaneous activity peaks of differing amplitudes.
Therefore, such cortical maps can represent superpositions of values with various amplitudes.

Topographic maps provide another point of contact between artificial neural networks and quantum computation, because the computational maps in the brain are large and dense enough that they can be usefully treated mathematically as fields, that is, as continuous distributions of continuous quantity \cite{FCNAI}. Such representations are suggestive of quantum mechanical wave functions, which are also continuous distributions of continuous quantity (the complex probability amplitude). In both cases these fields are treated mathematically as continuous functions on a continuous domain, and Hilbert spaces provide the mathematical framework for describing them \cite{FCFQIC}. In this paper we exploit this analogy to implement brain-inspired approaches to quantum machine learning.

Because of their spatial representation of values, topographic maps can be used to implement arbitrary functions in the brain, essentially by a kind of table lookup.
Suppose the brain needs to implement a (possibly nonlinear) transformation,
$\outval = f(\inval)$.
This can be accomplished by neural connections from locations $\loc=\map(\inval)$ in the input map to corresponding locations 
$\outloc=\map'(\outval)=\map'[f(\inval)]$
in the output map that represents the result.
Thus activity representing $\inval$ in the input map will cause corresponding activity representing $f(\inval)$ in the output map.
Moreover, a superposition of input values will lead to a corresponding superposition of output values.
Therefore, topographic representations allow the computation of nonlinear functions in linear superposition
\cite{Eliasmith2012,FCMC,FCNAI,FCNAI-ECSS,PAC},
which suggests their usefulness in quantum computation
\cite{FCFQIC}.
On the other hand, topographic maps make relatively inefficient use of representational resources, because every represented value has to have a location in the map.
(although this can be mitigated by means of \emph{coarse coding}, by which precise values are represented by a population of broadly tuned neurons with overlapping receptive fields \cite[pp.\ 91--96]{PDP1}\cite{Sanger1996}). 
Therefore, use of topographic representations will require a reasonably scalable quantum computing technology.
In this paper we explore topographic approaches to quantum computation with a focus on machine learning.

\section{Topographic Basis Maps}
In the brain, the state of a topographic map is a real-valued function defined over a (typically two-dimensional) space $\arbdom$.
To apply these ideas in quantum computation, we consider a quantum state $\Ket{\psi}$ in which the probability amplitude $\psi(\loc)$ at location $\loc \in \arbdom$ represents the value $\inval \in \invalsp$ via the correspondence $\loc = \map(\inval)$.
Here $\loc \in \arbdom$ may be a continuous index representing, for example, spatial location,
or a discrete quantum state, such as a wavelength or the state of a qubit register.
The states $\Ket{\loc}$ form a discrete basis or continuous pseudo-basis for the input and output quantum states.
In the continuous case, the input value $\inval$ is represented by a state $\Ket{\loc}$,
which is a Dirac unit impulse at $\loc = \map(\inval)$: that is, $\Ket{\loc} = \delta_\loc$ 
where $\delta_\loc(\altloc) = \delta(\altloc-\loc)$.
Similarly an output value $\outval$ is represented by the continuous basis state
$\Ket{\map'(\outval)} = \Ket{\map'[f(\inval)]} = \delta_{\map'(\outval)}$.
The states $\Ket{\loc}$ (for $\loc \in \arbdom$) form a continuous basis for the input and output quantum states.
We call such a representation (whether discrete or continuous) a \emph{topographic basis map}.

For such a continuous basis we can define a Hilbert-Schmidt linear operator
\[
 \gmap_f = \int_\invalsp \dif\inval~  \KetBra{\map'[f(\inval)]}{\map(\inval)} ,
\]
where $\invalsp$ is the space of input values.
(We write $\gmap = \gmap_f$ when $f$ is clear from context.)
This operator has the desired behavior:
$\gmap\Ket{\map(\inval)} = \Ket{\map'[f(\inval)]}$ for all $\inval \in \invalsp$.
In this manner the linear operator $\gmap$ computes the nonlinear function $f$ via the computational maps.
We call such an operator a \emph{graph kernel} because it uses the explicit \emph{graph} of $f$
[that is, the set of pairs $(\map'[f(\inval)], \map(\inval))$ for $\inval \in \invalsp$]
to do a kind of table lookup.\footnote{
It is not the same as the graph kernels used in machine learning applied to graph theory.
}

Notice that if the input map is a superposition of input values,
$\Ket{\psi} = a\Ket{\map(\inval)} + b\Ket{\map(\inval')}$,
then the output map will be a superposition of the corresponding results:
$\gmap\Ket{\psi} = a\Ket{\map'[f(\inval)]} + b\Ket{\map'[f(\inval')]}$.
Therefore, continuous topographic basis maps permit nonlinear functions to be computed in linear superposition (quantum parallelism). This is a step toward quantum computation, but $\gmap$ might not be unitary, and we have more work to do.

The reader might question the use of a continuous basis.
First, note that for separable Hilbert spaces, the continuous basis can always be replaced by an infinite discrete basis,
for example, by a discrete series of sinusoids or complex exponentials of the appropriate dimension. Second, the infinite discrete basis can be approximated by a finite discrete basis, 
for example, by band-limited sinusoids or complex exponentials. Such an approximation is especially appropriate for neural network machine learning, which requires only low-precision calculation.

%%%%%%%%%%
\subsection{Bijection}
We proceed to show several examples of nonlinear computations performed via topographic basis maps, beginning with a simple case and proceeding to more complex ones.
For simplicity, we will ignore the map $\map$ and consider computations from one quantum state to another.
We consider both 
one-dimensional continuous domains,
$\arbdom = [x_l, x_u]$,
and discrete domains, $\arbdom = \{x_1, \ldots, x_n\}$.
Typically the values would evenly spaced,
for example,
$\arbdom = \{0, \Delta x, 2\Delta x, \ldots, (n-1)\Delta x\}$,
but this is not required, and other spacings, such as logarithmic, might be useful.
(Logarithmic maps are found in some sensory cortical regions.)
In both the continuous and discrete cases the vectors 
$\{\Ket{x} \mid x \in \arbdom\}$ are an orthonormal basis
(composed of unit vectors in $\Complex^n$ for the discrete case 
and of delta functions in  $\Eltwo(\arbdom)$ for the continuous case).
For example, in the discrete case, the values $x_1, \ldots, x_n$ might be represented by the composite state of an $N$-qubit register, where $n=2^N$.
This might seem to require a large number of qubits, but even in the absence of coarse coding, seven qubits would be sufficient to represent values with 1\% precision, which is adequate for many machine learning applications.

We begin with a bijective scalar function $f:[-1, 1] \to [-1, 1]$.
The hyperbolic tangent (appropriately restricted\footnote{
For example, $f(x) = [\tanh(x)/\tanh(1)] |_{[-1,1]}$.
}), which is a useful sigmoid function for neural computation, is an example of such a function.
The graph kernel to compute the function topographically is
$\gmap = \intdom\dif x~ \KetBra{f(x)}{x}$.
Since $f$ is bijective, the adjoint of $\gmap$ is
\[
 \conjtrans{\gmap} = \intdom\dif x~ \KetBra{x}{f(x)} = \intdom\dif y~ \KetBra{f^{-1}(y)}{y} .
\]
It is easy then to see that $\gmap$ is unitary.
In general, we have:
\begin{prop}\label{prop:bij-unitary}
The graph kernel $\gmap$ of a continuous bijection $f: \arbdom \to \altdom$ is unitary.
\end{prop}
\begin{prf}
Substituting the above equations, observe:
\bea
 \conjtrans{\gmap}\gmap &=& 
    \left( \intalt \KetBra{f^{-1}(y)}{y} \dif y\right)
    \left( \intdom \KetBra{f(x)}{x} \dif x\right)\\
  &=& \intdom\intalt \Ket{f^{-1}(y)} \BraKet{y}{f(x)} \Bra{x}\ \dif y \dif x\\
  &=& \intdom\intalt \Ket{f^{-1}(y)} \BraKet{y}{f(x)} \dif y\ \Bra{x} \dif x\\
  &=& \intdom \KetBra{f^{-1}(f(x))}{x} \dif x\\
  &=& \intdom \KetBra{x}{x} \dif x\\
  &=& \I_\arbdom .
\eea
The fourth line follows from the ``sifting'' property of the Dirac delta.
Likewise,
\bea
 \gmap\conjtrans{\gmap} &=&
    \left( \intdom \KetBra{f(x)}{x} \dif x\right)
    \left( \intalt \KetBra{f^{-1}(y)}{y} \dif y\right)\\
  &=& \intalt\intdom \Ket{f(x)} \BraKet{x}{f^{-1}(y)} \Bra{y} \dif x \dif y\\
  &=& \intalt \KetBra{f(f^{-1}(y))}{y} \dif y = \I_\altdom .
\eea
Therefore $\gmap$ is unitary.
\end{prf}
In the discrete basis case, let $y_i = f(x_i)$; the unit vectors $\Ket{x_i}$ are a basis for $\Complex^n$,
and the unit vectors $\Ket{y_i}$  are also a (possibly different) basis for $\Complex^n$. 
The graph kernel is $\gmap = \sum_{i=1}^n \KetBra{y_i}{x_i} = \sum_{i=1}^n \KetBra{f(x_i)}{x_i}$,
which is also easily proved to be unitary.
More directly, $\gmap$ is a permutation matrix on the basis elements and therefore orthogonal.

If the input is a weighted superposition of values,
$\Ket{\psi} = \intdom\dif x\ p(x) \Ket{x}$,
then applying the kernel will give a corresponding superposition of the outputs:
$\gmap\Ket{\psi} = \intdom\dif x\ p(x) \Ket{f(x)}$.
The same applies, of course, in the discrete case.
Moreover, since the graph kernel is unitary,
the adjoint is the inverse:
if $\Ket{\phi} = \intalt\dif y\ q(y)\Ket{y}$,
then $\conjtrans{\gmap}\Ket{\phi} = \intalt\dif y\ q(y)\Ket{f^{-1}(y)}$.
That is, applying the adjoint to a superposition of outputs will compute a corresponding superposition of inputs.

%%%%%%%%
\subsection{Non-surjective injections}
As a further step towards the quantum computation of arbitrary functions by means of computational maps, we consider a relatively simple case: non-surjective injections (that is, one-to-one non-onto functions). 
We restrict our attention to finite domains and codomains. 
Therefore, 
let $\arbdom = \{x_1, \ldots, x_n\}$ and $\altdom = \{y_1, \ldots, y_m\}$,
where $n < m$,
and consider an injection $f: \arbdom \to \altdom$.
Input maps will be in an $n$-dimensional Hilbert space $\Hilb(\arbdom)$
and output maps will be in an $m$-dimensional Hilbert space $\Hilb(\altdom)$.
Since $n<m$, ancillary constants will need to be provided from a space $\HilbAnc$,
and so the complete input space will be in $\Hilb(\arbdom) \TP \HilbAnc$.
Our implementation will also generate ``garbage'' output in a space $\HilbGarb$,
and so the complete output space will be in $\Hilb(\altdom) \TP \HilbGarb$.
The input and output dimensions must be equal, and the simplest way to accomplish this is to make $\HilbAnc$ $m$-dimensional and $\HilbGarb$ $n$-dimensional,
so that our operator is an $mn$-dimensional Hilbert-space transformation.
Let $\{ \Ket{\exinbasis_1}, \ldots, \Ket{\exinbasis_m} \}$ be a basis for $\HilbAnc$
and let $\{ \Ket{\exoutbasis_1}, \ldots, \Ket{\exoutbasis_n} \}$ be a basis for $\HilbGarb$.
(We could in fact use the $\Hilb(\altdom)$ basis for $\HilbAnc$ and the $\Hilb(\arbdom)$ basis for $\HilbGarb$, but here we develop a more general result.)

Our goal will be to define a unitary $\Umap$ so that
$\Umap \Ket{x} \Ket{\exinbasis_1} = \Ket{f(x)} \Ket{\garbage}$,
where $\Ket{\exinbasis_1}$ is an ancillary constant and $\Ket{\garbage}$ is garbage.
As we will see, $\Umap$ can be implemented by an appropriate permutation of the input basis into the output basis, which can be expressed as the sum of several operators: 
$\Umap \eqdf \gmap + \nrmap + \Rmap + \Qmap$.
The work of the function $f$ is accomplished by the $\gmap$ component:
\[
 \gmap \eqdf \sum_{j=1}^n \Ket{f(x_j)} \Ket{\exoutbasis_1} \Bra{x_j} \Bra{\exinbasis_1} .
\]
Note that $\gmap \Ket{x_j} \Ket{\exinbasis_1} = \Ket{f(x_j)} \Ket{\exoutbasis_1}$,
and $\gmap$ is a bijection of the $n$-dimensional subspace 
$\Hilb(\arbdom) \TP \Hilb(\Ket{\exinbasis_1})$.
However $\gmap$ is not unitary since it is not a surjection.
See Figure \ref{fig-1}.

The $\nrmap$ component ensures that non-range elements of the codomain have preimages in the domain.
Therefore, let $\nonrange$ be the number of codomain elements that are not in the range of $f$,
that is, $\nonrange = | \altdom \setminus \Im f | = m-n$.
Call these non-range codomain elements
$\{ \nrbasis_1, \ldots, \nrbasis_\nonrange \} \subset \altdom$.
Then the $\nrmap$ component is defined:
\[
 \nrmap \eqdf
  \sum_{i=1}^\nonrange \Ket{\nrbasis_i} \Ket{\exoutbasis_1} \Bra{x_1} \Bra{\exinbasis_{i+1}} .
\]
Therefore, $\nrmap$ is a bijection of an $\nonrange$-dimensional subspace and each non-range element $\Ket{\nrbasis_i} \Ket{\exoutbasis_1}$
has a unique preimage $\Ket{x_1} \Ket{\exinbasis_{i+1}}$

%%%%%%%%%%%%%%
\Figure{fig-1}{Fig-1}{[scale=0.75]}{
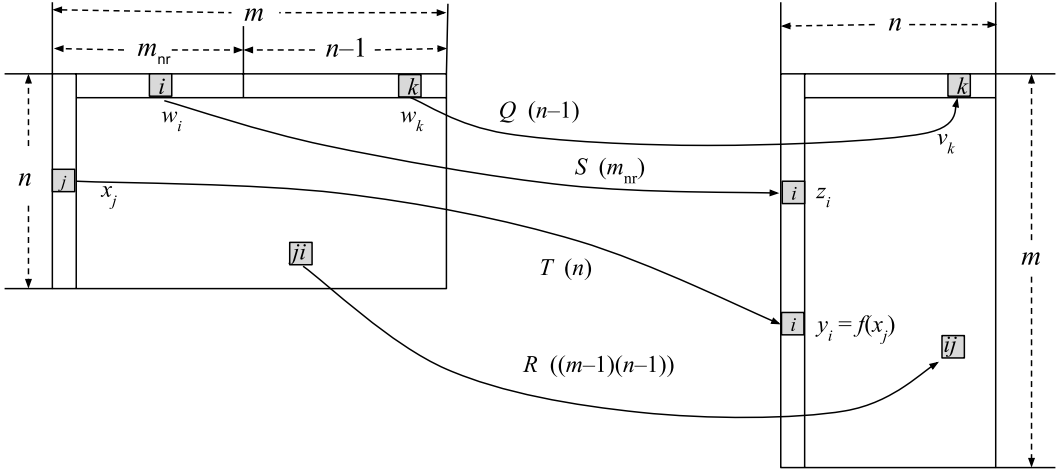
Permutation of basis vectors to implement non-surjective injections. After each component of the kernel, the number of basis vectors that it maps is indicated in parentheses; for example, $\Rmap$ maps $(m - 1)(n - 1)$ basis vectors. $\gmap$ maps function inputs to outputs, $\nrmap$ maps zero amplitudes to non-range codomain elements, and $\Rmap$ and $\Qmap$ bijectively map the remaining basis elements.
}
%%%%%%%%%%%%%%

Note that $\gmap$ transforms $n$ basis vectors (those for which the second register is 
$\Ket{\exinbasis_1}$), 
and $\nrmap$ transforms $\nonrange$ basis vectors (those for which the first register is $\Ket{x_1}$
and the second register is 
$ \Ket{\exinbasis_{2}}, \ldots,  \Ket{\exinbasis_{\nonrange+1}}$,
for a total of $n+\nonrange=m$ basis elements, but the input space has a total of $mn$ basis elements, and the remainder must be bijectively mapped. 

Therefore, to complete the unitary operator we add the following additional components:
\bea
 \Rmap &\eqdf&
   \sum_{i=2}^m \sum_{j=2}^n \Ket{y_i} \Ket{\exoutbasis_j} \Bra{x_j} \Bra{\exinbasis_i} , \\
 \Qmap &\eqdf& \sum_{j=1}^{n-1} \Ket{y_1} \Ket{\exoutbasis_{j+1}} \Bra{x_1} \Bra{\exinbasis_{\nonrange+j}} .
\eea
$\Rmap$ maps $(m-1)(n-1)$ basis elements: those for $i \ne 1$ and $j \ne1$, that is, those with neither $\Ket{x_1}$ in the first register nor $\Ket{\exinbasis_1}$ in the second. $\Qmap$ maps the remaining $n-1$ basis elements: those which have $\Ket{x_1}$ in the first register and 
$\Ket{\exinbasis_{\nonrange+1}}$ to $\Ket{\exinbasis_{\nonrange+n-1}}$
in the second (recall that $m=\nonrange+n$).

Notice that $\Umap$ maps every input basis vector into exactly one output basis vector and vice versa
(see Fig.\ \ref{fig-1}).
Summing the basis vector dyads for $\gmap, \nrmap, \Rmap, \Qmap$ gives:
\[
 n + \nonrange + (m-1)(n-1) + (n - 1)
 = m + (m-1)(n-1) + n - 1
 = mn .
\]
\begin{prop}\label{prop:nonsurjective-injection}
Let $\arbdom$ and $\altdom$ be finite sets with $n = | \arbdom |$, $m = | \altdom |$, and $m>n$.
Let $f: \arbdom \to \altdom$ be a non-surjective injection.
Let $\HilbAnc$ and $\HilbGarb$ be Hilbert spaces of dimension $m$ and $n$, respectively
(representing ancillary constant inputs and garbage outputs).
Let $\Ket{\Exinbasis}$ be a fixed basis vector of $\HilbAnc$
and $\Ket{\Exoutbasis}$ be a fixed basis vector of $\HilbGarb$.
Then there is a unitary operator
$\Umap \in \Lin[\Hilb(\arbdom) \TP \HilbAnc, \Hilb(\altdom) \TP \HilbGarb]$
such that for all $x \in \arbdom$,
\[
 \Umap (\Ket{x} \TP \Ket{\Exinbasis}) = \Ket{f(x)} \TP \Ket{\Exoutbasis} .
\] 
\end{prop}
\begin{prf}
The proposition follows from the construction preceding the proposition.
\end{prf}

If the input to $\Umap$ is a (normalized) superposition,
$\Ket{\psi} = \sum_k p_k \Ket{x_k}$,
then the output will be a normalized superposition of the corresponding function results:
$\Umap \Ket{\psi}\Ket{\Exinbasis} = \sum_k p_k \Ket{f(x_k)} \Ket{\Exoutbasis}$.

The dimension of the input and output spaces of this implementation is $mn$. A more resource efficient but also more complicated implementation operates on a space of dimension $\mathop{\rm LCM}(m,n)$.
The principle is the same: a permutation of the basis vectors.

%%%%%%%%%
\subsection{Non-injective surjections}
Next we consider functions $f: \arbdom \onto \altdom$ that are surjections but not injections; that is, $f$ maps onto $\altdom$ but might not be one-to-one.
This includes many useful functions, such as non-injective squashing functions and Gaussians, but also binary functions such as addition and multiplication (as explained later).

A non-injective function loses information, and thus it must be embedded in a larger injective function, which moreover must be unitary.
In particular, if $f$ is non-injective (e.g., $f(x)=f(x')$ for some $x \neq x'$), then the corresponding graph kernel will also be non-injective: $\gmap\Ket{x} = \Ket{y} = \gmap\Ket{x'}$ for $\Ket{x} \neq \Ket{x'}$.
Therefore $\gmap(\Ket{x} - \Ket{x'}) = \Zero$, which implies that $\Ket{x} - \Ket{x'}$ is in the null space of $\gmap$, $\nullsp(\gmap)$.
Therefore there is a bijection between the orthogonal complement of the null space,
$\nullsp(\gmap)\orcomp$, and the range of the operator, $\Im \gmap$.
Hence we can implement the non-injective operation by decomposing the input $\Ket{\psi}$ into orthogonal components $\Ket{\psi} = \Ket{\nonnpart}+\Ket{\nullpart}$,
where $\Ket{\nonnpart} \in \nullsp(\gmap)\orcomp$
and $\Ket{\nullpart} \in \nullsp(\gmap)$.
The $\Ket{\nonnpart}$ component is sufficient to determine the output, so
there is a bijection $\Ket{\nonnpart} \oneone \gmap\Ket{\psi}$,
and the $\Ket{\nullpart}$ component preserves the information to differentiate the inputs that map to this output.

To explain how this separation can be accomplished, we consider the finite-dimensional case, but it is easily extended.
Let $\arbdom = \{x_1, \ldots, x_n\}$ and $\altdom = \{y_1, \ldots, y_m\}$;
since $f$ is surjective, $m \leq n$.

The desired operator $\gmap \in \Lin(\Hilb,\Hilb')$,
where $\Hilb = \Hilb(\arbdom)$ is an $n$-dimensional Hilbert space with basis
$\{ \Ket{x_1}, \ldots, \Ket{x_n} \}$. 
The output space $\Hilb'$ is also $n$-dimensional, and $m$ of its basis vectors
$\Ket{y_1}, \ldots, \Ket{y_m}$ are used to represent a topographic map of the function's codomain (and range),
$\Im f = \altdom = \{y_1, \ldots, y_m\}$.
Therefore $\Hilb(\altdom)$ is a subspace of $\Hilb'$.
Let $\{ \Ket{\nullbasism_1}, \ldots, \Ket{\nullbasism_{n-m}} \}$
be a basis for $\Hilb(\altdom)\orcomp$, the orthogonal complement of $\Hilb(\altdom)$ in $\Hilb'$.
(This subspace will represent ``garbage'' with no computational relevance.)

We will define $\Ket{\nonnbasis_1}, \ldots \Ket{\nonnbasis_m}$ to be an orthonormal (ON) basis for 
$\nullsp(\gmap)\orcomp$
(the row space of $\gmap$), where $m$ is the rank of $\gmap$;
and we will define $\Ket{\nullbasis_1}, \ldots, \Ket{\nullbasis_\nullity}$ to be an ON basis for $\nullsp(\gmap)$,
where $\nullity = n-m$ is the nullity of $T$.
These bases will determine the orthogonal components $\Ket{\nonnpart} \in \nullsp(\gmap)\orcomp$ and $\Ket{\nullpart} \in \nullsp(\gmap)$ into which any input is separated.

An example will make this clearer. Suppose 
$\arbdom = \{k \Delta x \mid -N<k<N\}$ and $\altdom = \{k \Delta x \mid 0 \leq k < N\}$.
Let $\mathop{\rm abs} : \arbdom \to \altdom$ be the absolute value function (a noninjective surjection between these sets). 
A basis for the nonnull space $\nullsp(\gmap)\orcomp$ comprises $\Ket{\nonnbasis_0} = \Ket{0}$ and the vectors $\Ket{\nonnbasis_k} = (\Ket{{-}k \Delta x} + \Ket{k \Delta x})/\sqrt{2}$
(for $k=1,…,N-1$). 
(Note that $\Ket{{-}k \Delta x}$ and $\Ket{k \Delta x}$ are orthogonal vectors for $k \neq 0$.) 
These $N$ basis vectors are in a one-to-one relation with the codomain elements 
$\Ket{k \Delta x}$ (for $k=0,\ldots,N-1$). 
The nullity is $\nullity = (2N-1)-N=N-1$
and the basis vectors of the null space are:
\[ \Ket{\nullbasis_k} = (\Ket{{-}k \Delta x} - \Ket{k \Delta x})/\sqrt{2}
 \ \ \ \ (\mbox{for\ } k=1, \ldots ,N-1).
\]

Projection onto this space keeps the information necessary to distinguish the specific preimage that maps to a given output. 
In this case, it remembers the sign of the input: note that 
$\BraKet{\nullbasis_k}{k \Delta x} = +1/\sqrt{2}$ and 
$\BraKet{\nullbasis_k}{-k \Delta x} = -1/\sqrt{2}$.
Therefore, for input $\Ket{k \Delta x}$, the orthogonal components are 
$\Ket{\nonnpart} = \Ket{\nonnbasis_k}/\sqrt{2}$ and 
$\Ket{\nullpart} = \Ket{\nullbasis_k} /\sqrt{2}$;
and for input $\Ket{{-}k \Delta x}$, 
they are $\Ket{\nonnpart} = \Ket{\nonnbasis_k}/\sqrt{2}$ and 
$\Ket{\nullpart} = -\Ket{\nullbasis_k} /\sqrt{2}$. 
This completes the example and we return to the construction for an arbitrary non-injective surjection.

For each $y_i \in \altdom$, let 
$\preimg{f}{y_i} \eqdf \{ x \mid f(x)=y_i \}$ 
be the inverse image of $y_i$;
these are disjoint subsets of the domain $\arbdom$
and correspond to orthogonal subspaces of $\Hilb$.
Let $\premult_i \eqdf |\preimg{f}{y_i}|$ be the \emph{preimage multiplicity} of $y_i$,
where $n = n_1 + \cdots + n_m$.
Because different $y_i \in \Im \gmap$ have different preimage multiplicities, 
it will be convenient to separate $\gmap$ into $m$ constant functions
$\gmap_i: \preimg{f}{y_i} \to \{y_i\}$.
Therefore let
\bn\label{eqn:Ti-def}
\gmap_i \eqdf \RSRni \sum_{x_j \in \preimg{f}{y_i}} \KetBra{y_i}{x_j}
 = \Ket{y_i} \RSRni \sum_{x_j \in \preimg{f}{y_i}} \Bra{x_j}
 = \KetBra{y_i}{\nonnbasis_i} ,
\en
where we define the normalized basis vectors of $\nullsp(\gmap)\orcomp$:
\bn\label{eqn:ui-def}
 \Ket{\nonnbasis_i} \eqdf \RSRni \sum_{x_j \in \preimg{f}{y_i}} \Ket{x_j} ,
 i = 1, \ldots, m .
\en
The constant maps $\gmap_i$ operate independently on orthogonal subspaces of $\Hilb(\arbdom)$.

Note that $\BraKet{\nonnbasis_i}{x_j} \neq 0$ if and only if $y_i = f(x_j)$,
and in this case $\gmap_i\Ket{x_j} = \RSRni \Ket{y_i}$.
(The $1 / \sqrt{\premult_i}$ factor is required for normalization of the $\Ket{\nonnbasis_i}$.)
Clearly $\{\Ket{\nonnbasis_1}, \ldots \Ket{\nonnbasis_m}\}$ is an ON set,
since its elements are normalized linear combinations of disjoint sets of the basis vectors.
Therefore there is a one-to-one correspondence between the output vectors $\Ket{y_i}$ and the basis vectors $\Ket{\nonnbasis_i}$.

Next we characterize $\nullsp(\gmap)$ and $\nullsp(\gmap)\orcomp$.
Observe that $\Ket{\psi} \in \nullsp(\gmap_i)$ if and only if 
$\Zero = \gmap_i\Ket{\psi} = \Ket{y_i} \BraKet{\nonnbasis_i}{\psi}$,
that is, if and only if
$\BraKet{\nonnbasis_i}{\psi} = 0$.
Therefore, $\nullsp(\gmap_i)$ is the $n-1$ dimensional subspace orthogonal to $\Ket{\nonnbasis_i}$,
and $\nullsp(\gmap_i)\orcomp$ is the one-dimensional subspace spanned by $\{\Ket{\nonnbasis_i}\}$.
Therefore $\nullsp(\gmap)\orcomp$ is spanned by 
$\{\Ket{\nonnbasis_1}, \ldots, \Ket{\nonnbasis_m}\}$.

The operation $\gmap$ is implemented in two parts, one that handles the components representing the null space and the other that handles the components representing its orthogonal complement, the nonnull space.
The first part generates ``garbage,'' but is required for the operation to be invertible and hence unitary; the second part does the work of computing $f$.

We have $\Ket{\nullbasis_j} \in \Hilb$, the basis vectors for the $\nullsp(\gmap)$ subspace,
which has dimension $\nullity = n-m$.
Let $\{\Ket{\nullbasism_j} \mid j=1, \ldots \nullity \}$ be any basis for $\Hilb(\altdom)\orcomp$, the orthogonal complement of $\Hilb(\altdom)$ in $\Hilb'$, which also has dimension $n-m$.
We define
$\gmapnull \in \Lin(\Hilb,\Hilb')$
to map the nullspace components down to this $\nullity$-dimensional space:
\[ \gmapnull = \sum_{j=1}^\nullity \KetBra{\nullbasism_j}{\nullbasis_j} . \]

Since there is a one-one correspondence between  basis vectors $\Ket{\nonnbasis_i}$ and output vectors $\Ket{y_i}$,
we implement the function $f$ by an operator
$\gmapbij \in \Lin(\Hilb,\Hilb')$
defined
\[ \gmapbij = \sum_{i=1}^m \KetBra{y_i}{\nonnbasis_i} . \]
This maps the $n$-dimensional input into an $m$-dimensional output subspace.
As a result, the operator 
$\gmap \eqdf \gmapbij + \gmapnull \in \Lin(\Hilb,\Hilb')$ 
maps an input $\Ket{x}$ to the correct output $\Ket{f(x)}$, but with a scale factor and additional ``garbage.''
Specifically, for $x_j \in \preimg{f}{y_i}$,
\[
 \gmap \Ket{x_j}
  = (\gmapbij + \gmapnull) \Ket{x_j}
  =  \gmapbij \Ket{x_j} + \gmapnull \Ket{x_j} 
  = \RSRni \Ket{y_i} +  \Ket{\garbage_j} .
\]
where $\Ket{\garbage_j} = \gmapnull \Ket{x_j}$
and $\norm{\Ket{\garbage_j}} = \sqrt{\frac{n_i-1}{n_i}}$.
Note that the garbage $\Ket{\garbage_j}$ is superimposed on the desired output.
Subsequent computations operate on the 
$\Hilb(\altdom) = \{y_1, \ldots, y_m\}$ subspace and ignore the orthogonal subspace,
which contains the garbage (which nevertheless must be retained, since it is entangled with the computational results).

The $\gmap$ operator is just a transformation from the
$\{ \Ket{x_1}, \ldots, \Ket{x_n} \}$
basis to the
$\{ \Ket{y_1}, \ldots, \Ket{y_m}, \Ket{\nullbasism_1}, \ldots, \Ket{\nullbasism_{n-m}} \}$
basis, 
and is obviously unitary.
Therefore it can be approximated arbitrarily closely by a combination of 
$\H$ (Hadamard), $\CNOT$ (conditional NOT), and $\piby8$ ($\pi/8$) gates
\cite[\S 4.5]{Nielsen-Chuang}.

Unfortunately, the output vectors $\Ket{y_i}$ have amplitudes that depend on their preimage multiplicities.
That is, if $y_i = f(x_j)$, then $\gmap\Ket{x_j} = \RSRni \Ket{y_i} + \Ket{\garbage_j}$,
and we get different scale factors $\RSRni$ depending on the preimage multiplicity.
For $y_i = f(x_j)$, define $\csize_j \eqdf 1/\sqrt{\premult_i}$ to be this scale factor, so that
$\gmap \Ket{x_j} = \csize_j \Ket{f(x_j)} + \Ket{\garbage_j}$.
We would like to equalize the differing amplitudes but there does not seem to be unitary means for doing so.

It might seem that something like a Grover iteration \cite{Hoyer-rot} could be used to rotate the state vector from 
$\csize_j \Ket{y_i} + \Ket{\garbage_j}$ to $\Ket{y_i}$, but different $\csize_j$ require different numbers of iterations.
Something like Grover's algorithm with an unknown number of solutions could be used, but this would require trying multiple rotations.
Therefore, it seems better to accept the unwanted scale factors and work with them.
This means that any $\Ket{y_i}$ with positive amplitudes are considered outputs from the computation, and therefore all positive amplitudes are treated the same.

If we ignore the relative magnitudes of positive amplitudes, then a quantum state
$\Ket{\psi} = \sum_j p_j \Ket{x_j}$
(with $p_j \geq 0$)
can be interpreted as the set of all $x_j$ with positive amplitudes:
$\{x_j \in \arbdom \mid p_j > 0 \}$,
where we assume of course that $\sum_j p_j^2 \leq 1$.
Moreover, the sum can be strictly less than one only if there are additional ancillary states that make up the difference (like $\Ket{\garbage_j}$ in the previous example).
Applying $\gmap$ to such a state computes a state representing the image of the input set.
That is, $\gmap\Ket{\psi} = \sum_j p_j \csize_j \Ket{f(x_j)}$,
which represents the set
$\{f(x_j) \in \altdom \mid p_j > 0 \}$.
If $S \subseteq \arbdom$ is the set represented by $\Ket{\psi}$,
then $\gmap\Ket{\psi}$ represents its image $f[S]$.
Since zero amplitudes will always map to zero amplitudes and positive amplitudes will map to positive amplitudes, set membership will be appropriately mapped from the domain to the codomain.

\begin{prop}\label{prop:noninjective-surjection}
Let $\arbdom = \{ x_1, \ldots, x_n \}$ 
and $\altdom = \{ y_1, \ldots, y_m \}$ be finite sets with $m \leq n$.
Suppose $\{ \Ket{x_1}, \ldots, \Ket{x_n} \}$ is an ON basis for a Hilbert space $\Hilb$
and $\{ \Ket{y_1}, \ldots, \Ket{y_m} \}$ is an ON basis for a subspace $\Hilb(\altdom)$ of $\Hilb$.
For any surjection $f: \arbdom \onto \altdom$ there is a unitary operator $\gmap \in \Lin(\Hilb,\Hilb)$ such that
for any $x \in \arbdom$,
\[
 \gmap\Ket{x} = \frac{1}{\sqrt{\premult_x}} \Ket{f(x)} + \Ket{\garbage} ,
\]
\end{prop}
where $\premult_x = |\preimg{f}{f(x)}|$,
$\Ket{\garbage} \in \Hilb(\altdom)\orcomp$,
and $\norm{\,\Ket{\gamma}} = \sqrt{\frac{\premult_x-1}{\premult_x}}$.
\begin{prf}
As previously shown, this operator is given explicitly by
\[
 \gmap = \sum_{i=1}^m \KetBra{y_i}{\nonnbasis_i} + \sum_{k=1}^{n-m} \KetBra{\nullbasism_k}{\nullbasis_k} ,
\]
where $\Ket{\nonnbasis_i} = \RSRni \sum_{x \in \preimg{f}{y_i}} \Ket{x} $,
$\premult_i = |\preimg{f}{y_i}|$,
the $\Ket{\nullbasis_k}$ are an ON basis for the orthogonal complement of the space spanned by the 
$\Ket{\nonnbasis_i}$,
and the $\Ket{\nullbasism_k}$ are an ON basis for $\Hilb(\altdom)\orcomp$.
\end{prf}

%%%%%%%%%%%%%%%%%%%
\subsection{Arbitrary Functions}
\def\extdom{\Omega\degree}
\def\extcodom{\Omega^*}
\def\extin{x_0}
\def\extout{y_0}
\def\exbasis{\nullbasism}
\def\extinbasis{y}
\def\extoutbasis{x}
\def\logn{N}
\def\rangel{r}
\def\nrangel{z}
\def\nrangelv{{\bf z}}
\def\numcodom{n}
\def\numrange{{m_{\rm r}}}
\def\numnrange{{m_{\rm nr}}}
\def\ancnum{\numnrange}
\def\ancmax{m}
\def\ancmin{l}
In the preceding, we have assumed for convenience that the function is either non-injective or non-surjective, but not both.
The solutions are easily extended to arbitrary functions
since every function can be factored as a composition of an injection and a surjection.
More directly, we can combine Prop.\ \ref{prop:noninjective-surjection}, to implement the function as a surjection onto its range, 
with Prop.\ \ref{prop:nonsurjective-injection} to inject its range into its codomain.
Let the domain $\arbdom = \{ x_1, \ldots, x_n \}$, 
where $n = |\arbdom|$,
and let $\{ \Ket{x_1}, \ldots, \Ket{x_n} \}$ be the standard basis of $\Hilb(\arbdom)$.
Let $\extin \notin \arbdom$ be an additional value,
and define the extended domain $\extdom = \{ \extin \} \cup \arbdom$.
Then $\Hilb(\extdom)$ has basis 
$\{ \Ket{\extin}, \ldots, \Ket{x_n} \}$.
(For example, $\Hilb(\extdom)$ may be the state space of $\logn$ qubits,
where $n+1 = 2^\logn$.)
The additional $\Ket{\extin}$ dimension will carry the nullspace ``garbage'' from previous computations.
Similarly, let the codomain $\altdom = \{ y_1, \ldots, y_m \}$, 
where $m = |\altdom|$,
and let $\{ \Ket{y_1}, \ldots, \Ket{y_m} \}$ be the standard basis of $\Hilb(\altdom)$.
Let $\extout \notin \altdom$ be an additional value,
and define the extended domain $\extcodom = \{ \extout \} \cup \altdom$.
Then $\Hilb(\extcodom)$ has basis 
$\{ \Ket{\extout}, \ldots, \Ket{y_m} \}$.
The $\Ket{\extout}$ component carries the garbage in the output state.

Let $\{ \rangel_1, \ldots, \rangel_\numrange \} \eqdf \Im f$
and $\{ \nrangel_1 , \ldots, \nrangel_\numnrange \} \eqdf \arbdom \setminus \Im f$
be the range of $f$ and its complement, respectively;
$\numcodom = \numrange + \numnrange$.
As before, an $n$-dimensional input $\Ket{x_j}$ will be projected into orthogonal subspaces (the nonnull  and null spaces) of dimension 
$\numrange$ and $\nullity = n - \numrange$,
with basis vectors $\Ket{\nonnbasis_i}$ and $\Ket{\nullbasis_k}$, respectively.

An additional input quantum register will be used to provide the constant zero amplitudes for non-range elements for non-surjective functions.
The $m+1$ dimensional state of this register will be in, for convenience, 
$\HilbAnc = \Hilb(\extcodom)$
with basis $\{ \Ket{\extout}, \ldots, \Ket{y_m} \}$.
There will also be an additional output quantum register to hold the null space garbage for non-injective functions.
Its $n+1$ dimensional state is in, for convenience, 
$\HilbGarb = \Hilb(\extdom)$
with basis $\{ \Ket{\extin}, \ldots, \Ket{x_n} \}$.
Note that both the input and output spaces have dimension $(m+1)(n+1)$.
This is because the ancillary input register is in the same space as the regular output register, and the ancillary output register is in the same space as the regular input register. This can be confusing because, as will be seen, we use the extra \emph{output} vector $\Ket{\extout}$  as a constant in the ancillary \emph{input} register, and the extra \emph{input} vector $\Ket{\extin}$ appears in the ancillary \emph{output} register.

Our goal is to define unitary 
$\Umap \in \Lin[\Hilb(\extdom) \TP \HilbAnc, \Hilb(\extcodom) \TP \HilbGarb]$
so that
\[
 \Umap [ (\csize\Ket{x_j} + \dsize\Ket{\extin}) \TP \Ket{\extout}]
 = (\csize' \Ket{f(x_j)} + \dsize' \Ket{\extin}) \TP \Ket{\garbage} ,
\]
for scalars $\csize, \csize', \dsize, \dsize'$
and for $\Ket{\garbage} \in \HilbGarb$.
That is, the input register is initialized to the input $\Ket{x_j}$ with some positive amplitude $\csize$, possibly with superimposed garbage with amplitude $\dsize$; the ancillary input register is initialized to constant $\Ket{\extout}$. After computation, the output register will contain the function’s value $ \Ket{f(x_j)}$ with some positive amplitude $\csize'$; and superimposed garbage with amplitude $\dsize'$. The ancillary output register may also contain garbage. 
In other words, the argument of $f$ is in the first input register [corresponding to $\Hilb(\extdom)$], and its result is in the first output register [corresponding to $\Hilb(\extcodom)$], 
possibly with garbage in both its input and output $\Ket{\extin}$ components.
The input ancillary register is initialized to a constant $\Ket{\extout}$.

%%%%%%%%
\Figure{fig-2}{Fig-2}{[scale=0.75]}{
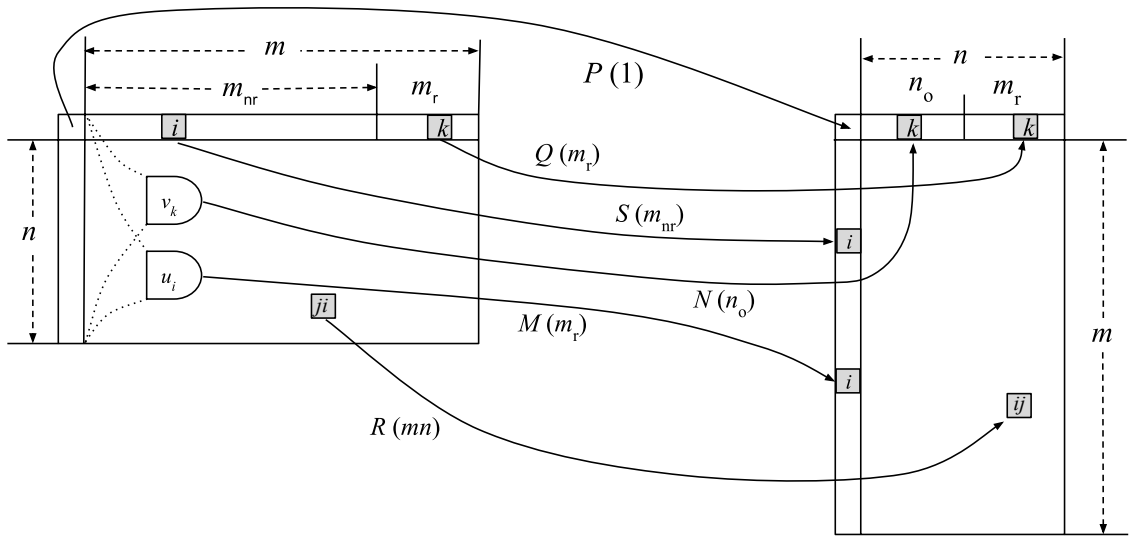
Permutation of basis vectors to implement arbitrary function. After
each component of the kernel, the number of basis vectors that it maps is
indicated in parentheses. For example, $\gmapbij$ maps $mn$ basis vectors. The shapes
labeled $\nonnbasis_i$ and $\nullbasis_k$ represent projection onto the basis vectors of the nonnull 
and null spaces, respectively.
}
%%%%%%%%

The work of computing $f$ is done by the graph kernel $\gmapbij$, which will map the $\Ket{\nonnbasis_i} \Ket{\extout}$ vectors into corresponding 
$(m+1)(n+1)$ dimensional output vectors 
$\Ket{\rangel_i}\Ket{\extin}$ in the output space $\Hilb(\extdom) \TP \HilbGarb$
(see Fig.\ \ref{fig-2}).
(The ancillary $\Ket{\extin} \in \HilbGarb$ output is required so that the input and output spaces have the same dimension.) 
To accomplish this mapping, define $\gmapbij$ as follows:
\[
 \gmapbij \eqdf \sum_{i=1}^\numrange
  \KetBra{\rangel_i, \extin}{\nonnbasis_i, \extout} .
\]
It maps $\numrange$ of the basis vectors of $\Hilb(\extdom) \TP \HilbAnc$ into $\numrange$ of the basis vectors of  $\Hilb(\extcodom) \TP \HilbGarb$ (see Fig.\ \ref{fig-2}). 
Specifically it is a bijection between the nonnull input subspace of $\Hilb(\extdom) \TP \Hilb(\{\extout\})$ and the range subspace of $\Hilb(\extcodom) \TP \Hilb(\{\extin\})$.

Another component of the transform will map the $\nullity$ null space components $\Ket{\nullbasis_k}\Ket{\extout}$
of the $m+1$ dimensional $\Ket{\extout}$ subspace of the input space:
\[ 
 \gmapnull \eqdf 
  \sum_{k=1}^{\nullity} \KetBra{\extout, x_k}{\nullbasis_k, \extout} .
\]
It maps them to $\nullity$ of the basis vectors $\Ket{\extout, x_k}$ of the $m+1$ dimensional $\Ket{\extout}$ subspace of the output space. $\gmapbij$ and $\gmapnull$ together handle non-injective functions.

For non-surjective functions, zero amplitudes are copied from the ancillary register $\Ket{\extin}\Ket{\extinbasis_i}$
into the appropriate non-range codomain components $\Ket{\nrbasis_i}\Ket{\extin}$:
\[
 \nrmap \eqdf \sum_{i=1}^\nonrange \KetBra{\nrbasis_i , \extin}{\extin , \extinbasis_i} .
\]
This is a mapping of $\nonrange$ basis vectors between the $\Ket{\extin}$ subspaces of the input and output spaces. 
The three operators $\gmapbij, \gmapnull, \nrmap$ handle the mapping of $f$ (and disposal of the null space).

We have to be careful, however, to handle all the $\Ket{\extin}\Ket{y_i}$
basis vectors since $\nonrange$ might be less than $n$ (see Fig.\ \ref{fig-2}). 
We map the remaining basis vectors of the 
$\Ket{\extin}$ subspace that were not used in the $\nrmap$ map to components of the $\Ket{\extout}$ subspace that are unfilled by $\gmapnull$:
\[
 \Qmap \eqdf \sum_{k=1}^\numrange \KetBra{\extout, x_{\nullity+k}}{\extin, y_{\numnrange+k}} .
\]
Note that $\nullity + \numrange = n = \numnrange + \numrange$,
so that all these vectors are bijectively mapped.

It remains to handle the other components of the input space in a unitary way.
The preceding maps have either $\Ket{\extin}$ or $\Ket{\extout}$, but not both, in the input register.
The $\Rmap$ operator maps the $mn$ basis vectors with neither:
$\Ket{x_j}\Ket{\extinbasis_i}$, for $i, j \neq 0$, map into their reverses:
\[
 \Rmap \eqdf \sum_{i=1}^m \sum_{j=1}^n \KetBra{y_i , \extoutbasis_j}{x_j , \extinbasis_i} .
\]
The state $\Ket{\extin}\Ket{\extout}$ remains, and it maps to its reverse:
\[
 \Pmap \eqdf \KetBra{\extout , \extin }{\extin , \extout} .
\]
In summary, $\gmapbij$ maps $\numrange$ basis vectors,
$\gmapnull$ maps $\nullity$ basis vectors,
$\nrmap$ maps $\nonrange$,
$\Rmap$ maps $mn$,
$\Qmap$ maps $\numrange$,
and $\Pmap$ maps one basis vector,
which accounts for all of the $(n+1)(m+1)$ basis vectors:
\bea
 \numrange + \nullity + \numnrange + mn + \numrange + 1
 &=& (\numrange + \nullity) + (\numnrange + \numrange) + mn  + 1\\
 &=& n + m + mn + 1\\
 &=& (m+1)(n+1).
\eea
The unitary operator to compute $f$ is the sum of these linear maps:
\[
 \Umap \eqdf \gmapbij + \gmapnull + \nrmap + \Rmap + \Qmap + \Pmap .
\]
\begin{prop}
Suppose $f: \arbdom \to \altdom$, where
$\arbdom = \{x_1, \ldots, x_n\}$
and $\altdom = \{y_1, \ldots, y_m\}$.
Let $n = |\arbdom|$
and $m = |\altdom|$.
Let $\Hilb(\extdom)$ be an $n+1$ dimensional Hilbert space with basis
$\{ \Ket{\extin}, \ldots, \Ket{x_n} \}$,
and let $\Hilb(\extcodom)$ be an $m+1$ dimensional space with basis
$\{ \Ket{\extout}, \ldots, \Ket{y_m} \}$.
Then there a unitary operator
\[
 \Umap \in \Lin[\Hilb(\extdom) \TP \Hilb(\extcodom), 
 \Hilb(\extcodom) \TP \Hilb(\extdom)]
\]
so that
for scalars $\csize, \dsize$ (with $\abs{\csize}^2 + \abs{\dsize}^2 = 1$)
and $x \in \arbdom$:
\[
 \Umap [ (\csize\Ket{x} + \dsize\Ket{\extin}) \TP \Ket{\extout}]
 = \csize\csize' \Ket{f(x), \extin} + \dsize' \Ket{\extout, \garbage} ,
\]
where $\csize = 1 / \sqrt{\premult_x}$,
where $\premult_x =  |\preimg{f}{f(x)}|$,
and $\abs{\csize\csize'}^2 + \abs{\dsize'}^2 = 1$.
\end{prop}
\begin{prf}
By construction we know:
\bea
 \Umap \Ket{x_j, \extout}
 &=& (\gmapbij + \gmapnull) \Ket{x_j, \extout} \\
 &=& \left( \sum_{i=1}^\numrange \KetBra{\rangel_i, \extin}{\nonnbasis_i, \extout} \right) \Ket{x_j, \extout}
  + \left( \sum_{k=1}^{\nullity} \KetBra{\extout, x_k}{\nullbasis_k, \extout} \right)  \Ket{x_j, \extout} \\
 &=& \Ket{ f(x_j), \extin } \BraKet{\nonnbasis_i}{x_j}
  + \Ket{\extout} \sum_{k=1}^{\nullity} \Ket{x_k}\BraKet{\nullbasis_k}{x_j} \\
 &=& \csize_j \Ket{ f(x_j), \extin } + \Ket{\extout, \garbage} ,
\eea
where 
$\Ket{\garbage} = (\sum_{k=1}^{\nullity} \KetBra{x_k}{\nullbasis_k})\Ket{x_j}$
and $\norm{\Ket{\extout, \garbage}} = \sqrt{\premult_j - 1} / \sqrt{\premult_j}$.
Furthermore, by construction,
\[
 \Umap \Ket{\extin, \extout}
 = \Pmap \Ket{\extin, \extout}
 = \Ket{\extout, \extin} .
\]
Therefore, in the general case where the input register is
$\csize\Ket{x_j} + \dsize\Ket{\extin}$
(with $\abs{\csize}^2 + \abs{\dsize}^2 = 1$)
we have:
\bea
 \Umap [ (\csize\Ket{x_j} + \dsize\Ket{\extin}) \TP \Ket{\extout}]
 &=& \csize\Umap \Ket{x_j, \extout} + \dsize\Umap \Ket{\extin, \extout} \\
 &=& \csize (\csize_j \Ket{ f(x_j), \extin } + \Ket{\extout, \garbage})
   + \dsize \Ket{\extout, \extin} \\
 &=& \csize \csize_j \Ket{ f(x_j), \extin }
   + \Ket{\extout} (\csize \Ket{\garbage} + \dsize \Ket{\extin}) .
\eea
\end{prf}
Therefore, the result that we want is in the first [$\Hilb(\extcodom)$] quantum register, but its $\Ket{\extout}$ component is garbage and should be ignored in subsequent computations.
Furthermore, the amplitude of desired result will decrease through successive computation stages through  attenuation by successive $1 / \sqrt{\premult_x}$ factors.

As discussed previously, quantum states with $\Ket{x_j}$ components with positive amplitudes represent sets of the corresponding $x_j$ ($j \neq 0$).
Applying $U$ to such a state yields a quantum state with positive amplitudes for the corresponding $\Ket{f(x_j)}$,
which represents the set of corresponding outputs $f(x_j)$.
%%%%%%%%
\section{Topographic Qubit Maps}
\subsection{Representation}
To further explore quantum computation by topographic maps, in this section we present an alternative representation of the maps and a circuit-based implementation of arbitrary functions on a finite domain.
Therefore, suppose $f: \arbdom \to \altdom$,
where the domain is $\arbdom = \{x_1, \ldots, x_n\}$
and the codomain is $\altdom = \{y_1, \ldots, y_m\}$.
In these \emph{topographic qubit maps}, each domain value $x_j$ or codomain value $y_i$ is assigned a separate qubit, whose state, for example,  $\Ket{\psi_j} = \qcomp{p_j} \Ket{0} + p_j \Ket{1}$,
where $\abs{\qcomp{p_j}}^2 + \abs{p_j}^2 = 1$,
represents the activity level of $x_j$ by the amplitude $p_j$.
This sort of one-out-of-$n$ representation might seem unrealistically inefficient, but
(1) we are assuming a scalable qubit implementation, which permits arrays of many thousands of qubits,
and (2) neural computations typically require only low precision (in the brain perhaps as little as one digit \cite[p.\ 378]{PDP2}). Therefore a quantity can be represented by a few tens of qubits.
In other words, our $m$ and $n$ will typically be small ($m, n < 100$).

Like the topographic basis maps, these topographic qubit maps can also be viewed as representations of subsets of the domain;
for $S \subseteq \arbdom$:
\[
 \Ket{S} = \sum_{x_j \in S} \Ket{1}_j + \sum_{x_j \notin S} \Ket{0}_j .
\]
That is, the $x_j$ qubit is in state $\Ket{1}$ if $x_j$ is in $S$ and is in state $\Ket{0}$ if it is not.
Therefore we use the notation $\singket{x_j}$ for the topographic map representing just the number $x_j$.
The set of representations of all possible subsets of $\arbdom$ is then an ON basis for the $2^n$-dimensional Hilbert space of these qubits.
The basis can be written:
\[
 \{ \Ket{k} \mid k \in \Two^n \} = \{ \Ket{S} \mid S \subseteq \Two^\arbdom \},
\]
where on the left $\Two^n$ is the set of $n$-bit binary strings,
and on the right $\Two^\arbdom$ is the powerset of $\arbdom$.
Therefore, the sets are basis states and as a consequence topographic qubit maps permit
\emph{multiple} sets to be processed in quantum superposition.
Moreover, 
because the sets are represented by computational basis vectors,
they can be copied without violating the no-cloning theorem.

By using amplitudes other than 0 and 1, we can represent fuzzy sets.
Suppose $S$ is a fuzzy set with membership function $\mu: S \to [0,1]$, and let $m_j = \mu(x_j)$.
Then $S$ is represented by the topographic qubit map
\[
 \Ket{S} = \sum_{j=1}^n m_j \Ket{1}_j + \sqrt{1-m_j^2}\ \Ket{0}_j .
\]
Fuzzy sets cannot, in general, be copied (nor can arbitrary superpositions of crisp sets).

With the topographic qubit representation, the transformations between computational maps will be implemented by the quantum circuit model, and so one might ask whether it would be simpler to implement ordinary binary digital quantum computation.
The answer is that computation on topographic maps can be implemented by a few relatively simple operations (described in the following subsections), so that computational maps buy a simpler quantum implementation at the cost of greater representational expense (number of qubits).
We expect topographic quantum computation to be more simply implemented than a full-scale digital quantum arithmetic-logic unit.
%%%%%%%%%%%%%%%
\subsection{Unary Functions}
An example will illustrate how to implement an arbitrary finite function $f: \arbdom \to \altdom$ by topographic qubit maps.
For any $y_i$ not in the range of $f$, we set its state $\Ket{\phi_i} = \Ket{0}$ supplied as an ancilla.
If $y_i$ is in the range of $f$, then it might be the image of a single domain element,
$x_j$, that is, $y_i = f(x_j)$, in which case we implement directly
$\Ket{\phi_i} = \Ket{\psi_j}$, transferring the state $\Ket{\psi_j}$ of input qubit $j$ to output qubit $i$.
If there are two domain values mapping into $y_i$,
say $f(x_j) = y_i = f(x_k)$, then we make $\Ket{\phi_i}$ the logical OR of $\Ket{\psi_j}$ and $\Ket{\psi_k}$.
This is accomplished by the two-input $\OR{2}$ gate:
\[
 \OR{2} \eqdf \CCNOT (\X \TP \X \TP \I),
\]
where $\CCNOT$ is the conditional-conditional-not or Toffoli gate.
The result of ORing the input states is:
\[
 \OR{2} (\Ket{\psi_j} \TP \Ket{\psi_k} \TP \Ket{1}) = \X\Ket{\psi_j} \TP \X\Ket{\psi_k} \TP \Ket{\phi_i} .
\]
The $\Ket{1}$ is an ancilla.
The result of the OR is in the third output qubit, and the first two output qubits, in which the negated inputs remain, are considered garbage. If 
$\Ket{\psi_j}=\qcomp{p}\Ket{0}+p\Ket{1}$ and $\Ket{\psi_k}=\qcomp{q}\Ket{0}+q\Ket{1}$,
then $\OR{2}$ transfers probability amplitudes as follows:
\bea
\OR{2}\Ket{\psi_j}\Ket{\psi_k}\Ket{1}
 &=& \OR{2}(\qcomp{p}\Ket{0}+p\Ket{1})(\qcomp{q}\Ket{0}+q\Ket{1})\Ket{1} \\
 &=& \qcomp{p}\qcomp{q}\OR{2}\Ket{001} + \qcomp{p}q\OR{2}\Ket{011}
       + p\qcomp{q}\OR{2}\Ket{101} + pq\OR{2}\Ket{111} \\
 &=& \qcomp{p}\qcomp{q}\Ket{110} + \qcomp{p}q\Ket{101} + p\qcomp{q}\Ket{011} + pq\Ket{001}.
\eea
Therefore, the third output qubit is the OR of the first two input qubits with the amplitudes shown.
If we interpret the squares of the amplitudes as probabilities, then $\OR{2}$ computes the correct probabilities for the third output qubit.
The first two output qubits are negated copies of the inputs, which are considered garbage but must be retained, for they are entangled with the third output.

If more than two domain values map into a single codomain value, then we use the multiple argument $\OR{n}$, which can be defined recursively in terms of $\OR{2}$:
\[
 \OR{n}\Ket{\psi_1} \cdots \Ket{\psi_n}\Ket{1}^{\TP (n-1)}
 \eqdf \OR{n-1}[(\OR{2}\Ket{\psi_1}\Ket{\psi_2}\Ket{1})
   \TP \Ket{\psi_3} \cdots \Ket{\psi_n} \Ket{1}^{\TP (n-2)}]\ \ \ (n>2).
\]
For completeness, we define $\OR{1} = \I$.

With the preceding motivation, we can give the construction for computing an arbitrary finite function by topographic qubit maps:
%%%%%%%%%
\def\zeroinput{z}
\def\oneinput{o}
\def\nroutput{c}
\def\bijinput{v}
\def\bijoutput{u}
\def\mulinput{x}
\def\muloutput{y}
\begin{prop}\label{prop:arbfinfunc}
Suppose $f: \arbdom \to \altdom$,
where $\arbdom = \{x_1, \ldots, x_n\}$
and $\altdom = \{y_1, \ldots, y_n\}$.
Let $\nonrange \eqdf n - | \Im f |$ be the number of codomain elements that are not in the range of $f$.
Let $\singval$ be the number of injective domain elements,
and let $\mulvalin = n-\singval$ be the number of non-injective domain elements.
Let $\mulvalout = | \Im f | - \singval$ be the number of non-injective range elements (i.e., those that are the image of two or more domain elements).
Then there is an $2\mulvalin + \singval + \nonrange - \mulvalout$ dimensional unitary operator $U_f$ that computes $f$ by topographic qubit maps:
\[
 U_f \singket{x_j} \Ket{0}^{\TP\nonrange} \Ket{1}^{\TP(\mulvalin-\mulvalout)}
 = \singket{y_i} \Ket{\garbage}  ,
\]
where $y_i = f(x_j)$
and $\Ket{\garbage}$ is $2(\mulvalin-\mulvalout)$ qubits of garbage.
\end{prop}
%%%%%
\begin{prf}
The inputs are the $n$ elements of the input map,
$\nonrange$ constant $\Ket{0}$ ancillae (for the non-range elements),
and $\mulvalin-\mulvalout$ constant $\Ket{1}$ ancillae for the $\OR{2}$ gates that map multiple domain elements to the same range element.
The latter is because each of the non-injective range elements requires a number of $\OR{2}$ gates that is one less than the number of its preimages; hence the $m_n$ non-injective range elements require $\mulvalin-\mulvalout$ $\OR{2}$ gates. 
Therefore, there are
\[
 n + \nonrange + \mulvalin-\mulvalout
 = (\singval + \mulvalin) + \nonrange + \mulvalin-\mulvalout
 = 2\mulvalin + \singval + \nonrange - \mulvalout
\]
input qubits.
The $\nonrange$ constant $\Ket{0}$s are passed directly to the output qubits for non-range codomain elements.
The $\singval$ qubits for injective inputs are passed to the same number of output qubits, permuted as required.
The outputs  of the ORs project onto the $\mulvalout$ qubits that represent range values with more than one pre-image.
Each $\OR{2}$ also generates two garbage qubits, for a total of $2(\mulvalin-\mulvalout)$.
Therefore the total number of output qubits is
\[
 \nonrange + \singval + \mulvalout + 2(\mulvalin-\mulvalout)
 = 2\mulvalin + \singval + \nonrange - \mulvalout,
\]
which is equal to the number of input qubits, as it should be.
Next we define $U_f$ explicitly as the tensor product of three operators:
\[ U_f \eqdf \Usingval \TP \Unonrange \TP \Umulval .\]
We will use the notation 
$\oket{q}$
to represent a $\Ket{1}$ state in qubit $q$, and 
$\zket{q}$
to represent a $\Ket{0}$ state in qubit $q$.

The $\Unonrange$ operator is an identity operation copying constant $\Ket{0}$ ancillae into the codomain elements that are not in the range of $f$.
Therefore, let $\{\nroutput_1, \ldots, \nroutput_\nonrange\} = \altdom - \Im f$ be this set,
and let $\Ket{\zeroinput_i}$ be ancillae qubits to provide constant $\Ket{0}$s.
Then $\Unonrange: \Hilb^\nonrange \to \Hilb^\nonrange$ is defined:
\[
 \Unonrange \eqdf \sum_{i=1}^\nonrange \zket{\nroutput_i} \zbra{\zeroinput_i}
  + \oket{\nroutput_i} \obra{\zeroinput_i} .
\]
That is, the states of the  $\zeroinput_i$ input qubits (intended to be $\Ket{0}$) are transferred to the $\nroutput_i$ output qubits.
This operator can be abbreviated by the following bracket notation:
\[
 \Unonrange \eqdf \qargsout{\nroutput_1,\ldots,\nroutput_\nonrange}
   \qargsin{\zeroinput_1,\ldots,\zeroinput_\nonrange} .
\]
It is just a permutation of the qubits, which might be implemented by SWAP operations.

The $\Usingval$ operator handles the domain elements that are mapped injectively to their images.
Therefore, let $\{\bijinput_1, \ldots, \bijinput_\singval\} \subseteq \Im f$ be the injective domain elements,
and let $\bijoutput_i = f(\bijinput_i)$ be the corresponding range elements.
Then $\Usingval: \Hilb^\singval \to \Hilb^\singval$ is a permutation of this subset of the topographic map elements:
\bea
  \Usingval \eqdf \qargsout{\bijoutput_1,\ldots,\bijoutput_\singval}
   \qargsin{\bijinput_1,\ldots,\bijinput_\singval} .
\eea

For $\Umulval$ we must OR together the domain elements corresponding to each range element with more than one preimage.
Therefore we define $\Umulval$ as a tensor product of operators for each such range element:
\[
 \Umulval \eqdf \bigTP_{i=1}^\mulvalout \Umulpart_i(\muloutput_i, \preimg{f}{\muloutput_i}) ,
\]
where these $\muloutput_i$ have more than one preimage element; for example,
$\preimg{f}{\muloutput_i} = \{\mulinput_1, \ldots, \mulinput_{\premult_i}\}$,
where $\premult_i = | \preimg{f}{\muloutput_i} | \geq 2$.
The output state $\Ket{\psi_i}$ for such a range element is the OR of the input states $\Ket{\xi_j}$ ($j = 1,\ldots \premult_i$) of its preimage elements:
\[
 \Ket{\psi_i} \Ket{\garbage} 
  = \OR{\premult_i} \Ket{\xi_1}\cdots\Ket{\xi_{\premult_i}} \Ket{1}^{\TP(\premult_i-1)} ,
\]
where $\OR{n_i }$ is a cascade of $n_i-1$ $\OR{2}$s and
$\Ket{\garbage}$ is $2\premult_i-2$ dimensional garbage.
This is accomplished by the operator
$\Umulpart_i(\muloutput_i,\{\mulinput_1, \ldots, \mulinput_{\premult_i}\})
\in \Lin(\Hilb^{2\premult_i-1} , \Hilb^{2\premult_i-1})$:
\[
 \Umulpart_i(\muloutput_i,\{\mulinput_1, \ldots, \mulinput_{\premult_i}\}) \eqdf
 \qargsout{ \muloutput_i, \garbage_1,\ldots,\garbage_{2\premult_i-2}} \OR{\premult_i}
  \qargsin{ \mulinput_1, \ldots, \mulinput_{\premult_i} , \oneinput_1, \ldots, \oneinput_{\premult_i-1} } ,
\]
where $\oneinput_1, \ldots, \oneinput_{\premult_i-1}$ are the qubits that provide ancillary $\Ket{1}$s for the ORs,
and the garbage outputs $\garbage_1,\ldots,\garbage_{2\premult_i-2}$ receive the negated inputs and intermediate OR outputs.The bracket notation identifies the qubits that are the inputs and outputs of $\OR{\premult_i}$.
This completes the construction of $U_f$.
\end{prf}
%%%%%%%%%
There are more efficient ways to compute $U_f$, but the above construction is easier to understand.

This basic approach can be used to approximate a variety of unary functions useful in neural networks, such as sigmoid functions, including non-injective, non-surjective squashing functions.
However, neural networks also require non-unary functions such as addition and multiplication, to which we now turn.

%%%%%%%%%%%%%%%%
\subsection{Binary Functions} 
In sensory cortical areas there are many topographic maps that represent two or more dimensions of a stimulus (e.g., retinal position and edge orientation); localized activity in these maps represent conjunctions of values on these dimensions.
Similarly, quantum computational maps can represent conjunctions of values as inputs or outputs of functions.

Suppose we want to compute a function $f: \arbdom \cross \arbdom \to \altdom$,
where $\arbdom = \{x_1, \ldots, x_n\}$
and $\altdom = \{y_1, \ldots, y_m\}$.
We will represent the input to the function by a two-dimensional array of qubits for each $(x_j,x_k)$ pair.
(They do not have to be physically arranged as a two-dimensional array so long as there is a qubit for each pair of values.)
This will require $n^2$ qubits, but we are assuming that $n$ is small because low precision is adequate for neural networks.
Therefore we expect the 2D map to comprise typically several thousand qubits.
The qubits representing the $(x_j,x_k)$ pairs are then mapped to the qubits representing the outputs $f(x_j,x_k)$ by the method described in Prop.\ \ref{prop:arbfinfunc}.

The $n^2$ conjunctions are computed by $n^2$ $\CCNOT$ gates, each of which requires a $\Ket{0}$ ancilla and generates two extra qubits (containing the inputs) in addition to the conjunction.
However, these extra qubits are passed along the rows and columns to be used in other conjunctions, and so there are only $2n$ total garbage qubits.
In summary, there are $2n+n^2$ input qubits (including $n^2$ ancillae)
and $n^2 + 2n$ output qubits (including $2n$ garbage).
That is, if $\Ket{\phi_j}$ is the state of element $j$ of one input map,
and $\Ket{\psi_k}$ is the state of element $k$ of the other input map,
then the state $\Ket{\chi_{jk}}$ of element $(j, k)$ of the two-dimensional map is computed by 
\[
 \Ket{\phi_j}\Ket{\psi_k}\Ket{\chi_{jk}} = \CCNOT \Ket{\phi_j}\Ket{\psi_k}\Ket{0}.
\]
If $\Ket{\phi_j} = \qcomp{p}\Ket{0} + p\Ket{1}$
and $\Ket{\psi_k} = \qcomp{q}\Ket{0} + q\Ket{1}$,
then 
\[
 \Ket{\phi_j}\Ket{\psi_k}\Ket{\chi_{jk}} = 
  \qcomp{p}\qcomp{q}\Ket{000} + p\qcomp{q}\Ket{100} + \qcomp{p}q\Ket{010} + pq\Ket{111} .
\]
The qubits are entangled, but the conjunction computes probability-like amplitudes if we interpret the squares of the amplitudes as probabilities.

Based on the foregoing, we define a unitary operator $\UOP$ on a $n^2+2n$ dimensional Hilbert space that does what amounts to an outer product on two one-dimensional maps to compute a two-dimensional map:
\[
 \Ket{\phi}\Ket{\psi}\Ket{\chi} = \UOP \Ket{\phi}\Ket{\psi}\Ket{0}^{\TP n^2},
\]
where $\Ket{\phi}$ and $\Ket{\psi}$ are $n$-dimensional,
and $\Ket{\chi}$ is $n^2$-dimensional.

To illustrate the use of computational maps to implement binary operations, we will use a simple, useful function, addition.
We want to define $\Sum: \arbdom\cross\arbdom \to \altdom$ so that
$\Sum(x,y) = x+y$,
but we have a problem, since the maximum value of $x+y$ is greater than the maximums of $x$ and $y$.
Since the range of numbers represented by our maps is quite limited, this is a more serious problem than overflow in binary addition.
One solution is to make the codomain map large enough;
for example, if $\arbdom = \{0, \Delta x, \ldots, (n-1)\Delta x\}$,
then let $\altdom = \{0, \Delta x, \ldots, 2(n-1)\Delta x\}$.
Generally, however, it is more convenient to have the codomain map be the same as the domain maps, since this facilitates composing functions.
Therefore, another solution is to scale either the inputs or the output so that we compute, for example,
$\mathop{\rm hsum}(x,y) = (x+y)/2$;
this is often useful if we know that we are going to scale the quantities anyway.
A third option is to compose the operator with squashing function, so that we compute, for example,
$\mathop{\rm tsum}(x,y) = \min(x+y, x_n)$, where $x_n = \max\arbdom$.
This is the solution that we will use for illustration.

If $\arbdom = \{0, \Delta x, \ldots, (n-1)\Delta x\}$, then the $(j,k)$ element of the two-dimensional qubit map will represent the pair of inputs $((j-1)\Delta x, (k-1)\Delta x)$.
This will be mapped to the sum $(j+k-2)\Delta x$ if $j+k-2 < n-1$,
and to the maximum value $(n-1)\Delta x$ otherwise.
Therefore the constant $j+k$ anti-diagonals above the $j+k-1 = n$ anti-diagonal each map to one value, $(j+k-2)\Delta x$, 
and all the elements below the $j+k-1 = n$ anti-diagonal map to $(n-1)\Delta x$.

Proposition \ref{prop:arbfinfunc} shows how to implement the truncated addition operation, but it treats it as a unary function on an $n^2$-dimensional space, which is wasteful since the intended output (the sum) is $n$-dimensional and the remaining $n^2-n$ elements are garbage. Therefore, we implement a unitary operator that directly maps the input pairs to the corresponding outputs.

To compute the outer product we require $n^2$ constant $\Ket{0}$ ancillae, and this computation also passes the two $n$-dimensional inputs through as garbage output. The qubit representing $(0, 0)$ maps bijectively to the output qubit $y_1$ representing 0. Each of the other $n-1$ output qubits $y_i$ ($i=2,\ldots,n$) has two or more domain pairs mapping to it. As before, let $n_i$ be the preimage multiplicity of output $i$ and note that $\sum_{i=1}^n n_i =n^2$. 
Each of these non-injective outputs receives its value from an $\OR{n_i }$ operation, which requires $n_i-1$ input $\Ket{1}$ ancillae and generates $2n_i-2$ qubits of output garbage ($n_i$ for the negated inputs and $n_i-1$ for the intermediate disjunctions). Therefore, the total number of $\Ket{1}$ ancillae is
\[
 \sum_{i=2}^n (n_i-1) = \sum_{i=2}^n n_i - (n-1) = (n^2-1)-n+1 = n^2-n .
\]
Moreover, the complete input dimension is $2n+n^2+n^2-n=2n^2+n$.
This is also the complete output dimension, for we have $n$ qubits for the function value, $2n$ qubits for the passed-through input arguments (garbage), and the garbage output from the OR gates, which is:
$\sum_{i=2}^n (2n_i-2) =2(n^2-n)$.
That is, the complete output dimension is $3n+2(n^2-n)=2n^2+n$.
In summary, there is a unitary operator $U_{\rm tsum} \in \Lin(\Hilb^{2n^2+n},\Hilb^{2n^2+n} )$ so that
\[
 U_{\rm tsum}\ \Ket{\{x\}}\ \Ket{\{y\}}\ \Ket{0}^{\TP n(n-1)}
 = \Ket{\{\mathop{\rm tsum}(x,y)\}}\ \Ket{\{x\}}\ \Ket{\{y\}}\ \Ket{\garbage} ,
\]
where the garbage $\Ket{\garbage}$ has dimension $2n(n-1)$ (the passed-through inputs may also be considered garbage).
Based on this example, we state a more general result.

\begin{prop}
Suppose $f: \arbdom \cross \arbdom \to \arbdom$ and let $n=|\arbdom|$. 
Let $\nonrange =n- |\Im f |$ be the number of codomain elements that are not in the range of $f$. 
Then there is a unitary operator $U_f \in \Lin(\Hilb,\Hilb)$, where $\Hilb$ is $2n^2+n+2\nonrange$ dimensional Hilbert space, such that:
\[
 U_f\ \singket{x}\ \singket{y}\ \Ket{0}^{\TP (n^2+\nonrange)}\ \Ket{1}^{\TP (n^2-n+\nonrange)}
  = \singket{f(x,y)}\ \singket{x}\ \singket{y}\ \Ket{\garbage} ,
\]
where the garbage $\Ket{\garbage}$ has dimension $2(n^2-n+\nonrange)$ (the $2n$ passed-through inputs may also be considered garbage).
\end{prop}
\begin{prf}
The operator is constructed very similarly to $U_{\rm tsum}$, but we also have to consider non-range codomain elements for non-surjective functions, which didn’t occur in that case.
As before, the computation of the outer product will require $n^2$ ancillary $\Ket{0}$ inputs and it will generate $2n$ qubits containing the passed-through inputs. 
We can consider disjoint subsets of the codomain. 
Codomain elements that are not in the range of $f$ will need to be sent a $\Ket{0}$ state, for which we need an additional $\nonrange$ ancillary $\Ket{0}$ inputs. 
Let $\singval$ be the number of input pairs that are mapped one-to-one to the corresponding outputs; they neither require ancillary constants nor generate garbage. 
The remaining codomain elements are range elements with $\premult_i \geq 2$; 
let $\mulvalout = n-\nonrange-\singval$ be the number of them. 
Each of these will receive the OR of the corresponding (preimage) domain elements. 
As we saw previously, the $\OR{\premult_i}$ operation requires $\premult_i-1$ ancillary $\Ket{1}$ qubits and produces $2\premult_i-2$ qubits of garbage. 
Therefore, the total number of $\Ket{1}$ qubits required for the $n^2-\singval$ input pairs mapping to the $\mulvalout$ non-injectively mapped range elements is:
\[
 \sum_{i=1}^\mulvalout (\premult_i - 1) = n^2 - \singval - \mulvalout = n^2 - n + \nonrange ,
\]
since $\mulvalout = n - \nonrange - \singval$. 
The garbage generated by the ORs is then 
$\sum_{i=1}^\mulvalout (2n_i - 2) = 2(n^2 - n + \nonrange)$.
The complete input dimension is $2n$ (arguments) + $(n^2+\nonrange)$ (for $\Ket{0}$ ancillae) + $(n^2-n+\nonrange)$ (for $\Ket{1}$ ancillae) = $2n^2+n+2\nonrange$. 
The complete output dimension is $3n$ (arguments and result) + $2(n^2-n+\nonrange)$ (garbage) = $2n^2+n+2\nonrange$.
\end{prf}
The same approach can be used for operations with more than two arguments, but the number of qubits increases exponentially with the number of arguments.

%%%%%%%%%
\section{Conversions Between Representations}
Ordinary binary representations can be translated to topographic qubit maps by a unitary demultiplexer $\Udemux$ that operates on an $m$-qubit binary number $\Ket{k}$ and directs a $\Ket{1}$ qubit to the $k$th of $n = 2^m$ output qubits (the remainder receiving $\Ket{0}$).
Let $\singket{k}$ be the resulting computational map.
Then:
\[  
 \Udemux \Ket{k} \Ket{1} \Ket{0}^{\TP (n-1)} = \Ket{k}\singket{k} .
\]
$\Udemux$ operates on an $m+n = m+2^m$ dimensional Hilbert space.
A demultiplexer can be implemented with $\CSWAP$ (Fredkin) gates \cite{Fredkin-Toffoli-82}.

The opposite translation, from a computational map to a binary representation, is more complicated.
First, we must decide what we want it to do, for in general a topographic qubit map represents multiple values with different amplitudes,
$\Ket{\psi} = \bigTP_{j=1}^n \qcomp{p_j} \Ket{0} + p_j \Ket{1}$,
where $\abs{\qcomp{p_j}}^2 + \abs{p_j}^2 = 1$.
Which $x_j$ should it produce?
The one with the maximum amplitude?
(And what if several have the same maximum amplitude?)
An $x_j$ chosen probabilistically based on $\abs{p_j}^2$?
The binary representation of a weighted average
$n^{-1}\sum_{j=1}^n p_j x_j$?
A normalized superposition of all the values?
The answer is not apparent, so we leave the question open.\footnote{
It is easy however to produce the binary representation of either the maximum or minimum number with unit amplitude ($p_j=1, p'_j=0$) in a map.
}

\section{Applications to Quantum Machine Learning}
Given this general ability to compute non-unitary and even nonlinear functions by means of topographic qubit maps, it is possible to do the operations useful for machine learning such as inner products and sigmoid nonlinearities.
For example, an inner product of $N$-dimensonal vectors requires $N$ multiplications and $N-1$ additions.
If $|\arbdom| = n$, then each multiplication and addition will require approximately $2n^2$ qubits,
for a total of about $2 N^2 n^2$.

For one layer of a neural network, say $N$ neurons projecting through an $M \cross N$ weight matrix into $M$ neurons, we must do $M$ inner products with the input.
Since crisp sets are represented by topographic qubit maps that are basis vectors in the computational basis, they can be copied.
Therefore, the $N$-dimensional input vector can be copied $M-1$ times to do the $M$ inner products.
(This requires $M-1$ $\CNOT$ gates and $(M-1)n$ ancillary $\Ket{0}$ qubits.
Overall, one layer requires about $2 M N^2 n^2$ qubits for the computation (not including ancillary qubits).
%%%%%%%%%%%%
\section{Conclusions}
Topographic (computational) maps are widely used in the brain to implement simultaneous nonlinear vector transformations on multiple inputs.
In this chapter we have explored two approaches to quantum topographic computing with a focus on brain-inspired machine learning applications.
The first, called a topographic basis map, assigns locations in the map to state vectors in a continuous or discrete basis for a quantum Hilbert space.
Arbitrary functions can be implemented on such maps, which can be interpreted as representing crisp sets of inputs, 
but there is an unavoidable data-dependent attenuation of the result (relative to a ``garbage'' state) that is not easily avoidable.
The second approach, called a topographic qubit map, assigns a separate qubit to each location in the map, and uses the relative amplitude of the $\Ket{1}$ and $\Ket{0}$ states to represent the presence of values in the (crisp or fuzzy) set represented by the map.
Arbitrary functions on these maps are implemented by well-known quantum logic gates.
In particular, computational maps enable the implementation of the functions commonly used in artificial neural networks.

%%%%%%%%%%%%
\newpage
\bibliographystyle{plain}
\bibliography{BJM2018,LNUCspecificplus,PhysRevA.62.052304}

\begin{thebibliography}{10}

\bibitem{Eliasmith2012}
C.~Eliasmith, T.~C. Stewart, X.~Choo, T.~Bekolay, T.~DeWolf, Y.~Tang, and
  D.~Rasmussen.
\newblock A large-scale model of the functioning brain.
\newblock {\em Science}, 338:1202--1205, 2012.

\bibitem{Fredkin-Toffoli-82}
E.~F. Fredkin and T.~Toffoli.
\newblock Conservative logic.
\newblock {\em Int. J. Theo. Phys.}, 21(3/4):219--253, 1982.

\bibitem{Furber-SpiNNaker}
E.~F. Furber, D.~R. Lester, L.~A. Plana, J.~D. Garside, E.~Painkras, S.~Temple,
  and A.~D. Brown.
\newblock Overview of the {SpiNNaker} system architecture.
\newblock {\em IEEE Transactions on Computers}, 62:2454--2467, 2013.

\bibitem{Hoyer-rot}
Peter H\o{}yer.
\newblock Arbitrary phases in quantum amplitude amplification.
\newblock {\em Phys. Rev. A}, 62:052304, Oct 2000.

\bibitem{FCMC}
Bruce~J. MacLennan.
\newblock Field computation in motor control.
\newblock In Pietro~G. Morasso and Vittorio Sanguineti, editors, {\em
  Self-Organization, Computational Maps and Motor Control}, pages 37--73.
  Elsevier, 1997.
\newblock Also available from {\tt web.eecs.utk.edu/\textasciitilde mclennan}.

\bibitem{FCNAI}
Bruce~J. MacLennan.
\newblock Field computation in natural and artificial intelligence.
\newblock {\em Information Sciences}, 119:73--89, 1999.
\newblock Also available from {\tt web.eecs.utk.edu/\textasciitilde mclennan}.

\bibitem{FCNAI-ECSS}
Bruce~J. MacLennan.
\newblock Field computation in natural and artificial intelligence.
\newblock In {R. A.} {Meyers et al.}, editor, {\em Encyclopedia of Complexity
  and System Science}, chapter 6, entry 199, pages 3334--3360. Springer, 2009.

\bibitem{PAC}
Bruce~J. MacLennan.
\newblock The promise of analog computation.
\newblock {\em International Journal of General Systems}, 43(7):682--696, 2014.

\bibitem{FCFQIC}
Bruce~J. MacLennan.
\newblock Field computation: A framework for quantum-inspired computing.
\newblock In Siddhartha Bhattacharyya, Ujjwal Maulik, and Paramartha Dutta,
  editors, {\em Quantum Inspired Computational Intelligence: Research and
  Applications}, chapter~3, pages 85--110. Morgan Kaufmann / Elsevier,
  Cambridge, MA, 2017.

\bibitem{PDP2}
J.~L. McClelland, D.~E. Rumelhart, and PDP~Research Group.
\newblock {\em Parallel Distributed Processing: Explorations in the
  Microstructure of Cognition, Volume 2: Psychological and Biological Models}.
\newblock MIT Press, Cambridge, 1986.

\bibitem{Morasso-Sanguineti-SOCMMC}
P.~G. Morasso and V.~Sanguineti.
\newblock {\em Self-Organization, Computational Maps and Motor Control}.
\newblock North-Holland, Amsterdam, 1997.

\bibitem{Neftci2013}
E.~Neftci, J.~Binas, U.~Rutishauser, E.~Chicca, G.~Indiveri, and R.~J. Douglas.
\newblock Synthesizing cognition in neuromorphic electronic systems.
\newblock {\em Proceedings of the National Academy of Sciences of the United
  States of America}, 110:E3468--E3476, 2013.

\bibitem{Nielsen-Chuang}
Michael~A. Nielsen and Isaac~L. Chuang.
\newblock {\em Quantum Computation and Quantum Information}.
\newblock Cambridge, Cambridge, 10th anniversary edition edition, 2010.

\bibitem{OReilly-CCN}
R.~C. O'Reilly, Y.~Munakata, M.~J. Frank, T.~E. Hazy, and contributors.
\newblock {\em Computational Cognitive Neuroscience}.
\newblock http://ccnbook.colorado.edu/, 2nd edition, 2014.

\bibitem{PDP1}
D.~E. Rumelhart, J.~L. McClelland, and PDP~Research Group.
\newblock {\em Parallel Distributed Processing: Explorations in the
  Microstructure of Cognition, Volume 1: Foundations}.
\newblock MIT Press, Cambridge, 1986.

\bibitem{Sanger1996}
T.~D. Sanger.
\newblock Probability density estimation for the interpretation of neural
  population codes.
\newblock {\em Journal of Neurophysiology}, 76:2790--2793, 1996.

\bibitem{Schuld-QQNN}
M.~Schuld, I.~Sinayskiy, and F.~Petruccione.
\newblock The quest for a quantum neural network.
\newblock {\em Quantum Information Processing}, 13:2567--2586, 2014.

\bibitem{Schuld-IQML}
M.~Schuld, I.~Sinayskiy, and F.~Petruccione.
\newblock An introduction to quantum machine learning.
\newblock {\em Contemporary Physics}, 56:172--185, 2015.

\end{thebibliography}
\end{document}